\documentclass[sigconf]{acmart-me}
\usepackage{booktabs}
\usepackage{url}
\usepackage{color}
\usepackage{enumitem}
\setcopyright{rightsretained}

\acmConference[MediaEval'21]{Multimedia Evaluation Workshop}{13-15 December 2021}{Bergen, Norway, and Online}
\acmYear{2021}
\copyrightyear{2021}

\begin{document}
\title{Predicting Media Memorability: Comparing Visual, Textual and Auditory Features}
\author{Lorin Sweeney, Graham Healy, Alan F. Smeaton}
\affiliation{Insight Centre for Data Analytics, Dublin City University, Glasnevin, Dublin 9, Ireland\\}
\email{lorin.sweeney8@mail.dcu.ie, graham.healy@dcu.ie, alan.smeaton@dcu.ie}

\renewcommand{\shortauthors}{L. Sweeney, G. Healy, A. Smeaton}
\renewcommand{\shorttitle}{Predicting Media Memorability}

\begin{abstract}
This paper describes our approach to the Predicting Media Memorability task in MediaEval 2021, which aims to address the question of media memorability by setting the task of automatically predicting video memorability. This year we tackle the task from a comparative standpoint, looking to gain deeper insights into each of three explored modalities, and using our results from last year's submission (2020) as a point of reference. Our best performing short-term memorability model (0.132) tested on the TRECVid2019 dataset---just like last year---was a frame based CNN that was not trained on any TRECVid data, and our best short-term memorability model (0.524) tested on the Memento10k dataset, was a Bayesian Ride Regressor fit with DenseNet121 visual features. 
    
\end{abstract}
\maketitle

\section{Introduction And Related Work}
\label{sec:intro}
In the ever expanding storm of social media, the need for tools that help us wade through daily digital torrents will only grow. It can be argued that memorability is a measure whose shape uniquely fits the jagged edged problem of media content curation. Our lack of meta-cognitive insight into what we will ultimately remember or forget \cite{photograph2013memorable}, casting clouds of obscuring cover over answers that will thread together our sense of self, is what motivates and brings meaning to the exploration of memorability---generally known as the likelihood of an observer remembering a repeated piece of media in a stream of media. 

This paper outlines our participation in the 2021 MediaEval Predicting Media Memorability Task \cite{ME2021}, which includes an extended subset of last year's TRECVid 2019 Video-to-Text dataset  \cite{mediaeval2020memory}, and Memento10k \cite{mem10k}---a large and diverse short-term video memorability dataset. This year, short-term (after minutes) memorability is further sub-categorised into \textit{raw} and \textit{normalised}, and long-term (after 24-72 hours) memorability is kept the same. Additionally, two video memorability prediction sub-tasks were put forward, the first (sub-task 1) following the standard train with provided data to generate predictions, and the second (sub-task 2) taking the form of a constrained generalisation task---where the training and testing data must be from different sources. Further information about the datasets, annotation protocol, pre-computed features, and ground-truth data can be found in the task overview paper \cite{ME2021}.

With last year's task \cite{mediaeval2020memory} including audio as part of the video data for the first time, its impact in the context of multi-modal media was thrust into the limelight. While no conclusive findings were established, the best long-term memorability prediction came from an xResNet34 trained purely on audio spectrograms \cite{sweeney2020leveraging}, suggesting that the audio modality provides a degree of useful information during video memorability recognition tasks. Additionally, follow-on work found evidence to suggest that ``audio plays a contextualising role, with the potential to act as a signal or a trigger that aids recognition'' depending on the extent of high-level human understandable information it contains, and the context in which it is presented \cite{sweeney2021audio}.

Many previous works have firmly established the utility of combining features from more than one modality, and highlighted the effectiveness of combining deep visual features in conjunction with semantically rich features, such as captions; emotions; or actions in order to predict media memorability \cite{azcona2019ensemble,mem10k,zhao2020multi, reboud2020predicting}. However, given that this year's sub-task 1 could be viewed as a natural extension of the previous year's task, that this year's sub-task 2 is a generalisation task, and that the previous years official results were abnormally low across the board, we opted treat this year's task as one of insight rather than optimisation, keeping modalities separate, rather than following state of the art by combining features across modalities---which ultimately obscures the extent to which each modality contributes to the final memorability score prediction---and limiting each of our runs to one modality.

\section{Approach}
\label{sec:approach}
Both datasets are comprised of three subsets, a training set; a development set; and a test set, with the TRECVid training set comprising of 588 videos, and Memento10k 7,000 videos. The development sets contain 1,116 and 1,500 videos respectively, and the test sets contain 500 and 1,500 videos respectively. Our approach this year was to use the task as an opportunity to compare our results from last year, cutting down the complexity and focusing on one of three modalities, visual, textual, and auditory.

\textbf{Visual:}
For our visual approach, we implemented two methods, the first of which was a Bayesian Ridge Regressor (BRR) that we fit with default sklearn \cite{pedregosa2011scikit} parameters using the provided DenseNet121 \cite{huang2019convolutional} features (which were extracted from the first, middle, and last video frames), and the second method was an ImageNet-pretrained xResNet50 that was either fine-tuned (for 50 epochs, with a maximum learning rate of 1e-3, and weight decay of 1e-2) on the Memento10k training data  and then further fine-tuned (for 10 epochs, with a maximum learning rate of 3e-2, and weight decay of 1e-1) on the TRECVid development set videos, fine-tuned on the Memento10k training data, or fine-tuned on the LaMem \cite{khosla2015understanding} dataset depending on the run and its restrictions.

\textbf{Textual:}
 For our textual approach, we implemented a caption model, the AWD-LSTM (ASGD Weight-Dropped LSTM) architecture \cite{AWD-LSTM}, a highly regularised and competitive language model. Transfer learning was used in order to fully avail of the high-level representations that a language model offers. The specific transfer learning method employed was UMLFiT \cite{UMLFit}, which uses discriminative fine-tuning, slanted triangular learning rates, and gradual unfreezing. The language model was pre-trained on the Wiki-103 dataset, and fine-tuned (for 10 epochs, with a maximum learning rate of 2e-3, a weight decay of 1e-2, and a dropout multiplier of 0.5) on the first 300,000 captions from Google's Conceptual Captions dataset \cite{GCC}. The encoder from that fine-tuned language model was then used in each of our caption models, which were either trained (for 15 epochs, with a maximum learning rate of 1e-3, a weight decay of 1e-2, and a dropout multiplier of 0.8) on a paragraph of all five Memento10k training captions, or additionally fine-tuned on the first TRECVid development set captions to predict memorability scores rather than the next word in a sentence.

\textbf{Auditory:}
Initially, we extracted Mel-frequency cepstral coefficients from the videos that had audio, stacked them together with their delta coefficients in order to create a three channel spectrogram images, and used them to train an ImageNet-pretrained xResNet34. However, after experimenting with VGGish \cite{vggish} features---extracting 128-dimensional embeddings for each second of the first three seconds of audio, resulting in a 384-dimensional feature set per video---and using them to fit a BRR, we noticed marginally, but consistently better results, and opted to use them in favour of spectrogram images in our final run submissions.

\section{Discussion and Outlook}
\label{sec:discussion}
Tables 1 and 2 show the Spearman scores (r\textsubscript{s}) and Pearson scores (r) for our runs from sub-task 1, with Table 1 showing the scores for our runs tested on the official TRECVid test set, and Table 2 showing the scores which came from the official Memento10k test set. Table 3 shows the r\textsubscript{s} and r scores for our runs from sub-task 2---the generalisation task---which were trained without any TRECVid data, and tested on the official TRECVid test set.

Although more TRECVid videos were provided this year compared to last year (1,704 vs 1,000), and even though our short-term TRECVid test scores (Table 1) are roughly double what they were last year, the scores are still quite low compared to expected results from training validation, and the long-term scores are shockingly poor. While it is not possible to pinpoint the exact cause of this, it is quite likely that either there is lack of distributional overlap between videos used to train and test our models, or that there still are not enough videos to be able to properly generalise. Both of these possibilities are supported by the fact that our best performing TRECVid run---just like last year---came from a model not trained on any TRECVid data, but purely on memento10k data (Table 3. \textit{xResNet50 Frames Memento}), which is a much larger, varied, and ``in-the-wild'' video memorability dataset than TRECVid.

Results from Table 2 show that the best performing model on the Memento10k dataset was a BRR fit on DenseNet121 features, indicating that visual features contribute quite a lot to the overall memorability of a video. The next best model was a BRR trained on VGGish audio features, which is very interesting as the Memento10k ground-truth scores were gathered with the videos being played without sound. The stark order of magnitude difference in performance between a BRR trained on Memento10k data (0.524) and one trained on TRECVid data (0.053), raises some interesting questions concerning the nature of the differences in visual content between these two datasets, which unfortunately cannot be answered in this paper.

Results from the generalisation task (Table 3) further highlight the aforementioned potential distributional problems with the TRECVid dataset. Given that the performance of both the frame based and caption based models is worse on the TRECVid test set when further fine-tuned on the TRECVid training and development data, a detailed exploration and investigation into the nature and distributions of the TRECVid subsets could be very fruitful.

While insights into possible causes of last year's uncharacteristically low task-wide scores across participant submissions were gained, few tangible insights into the influence of each of the explored modalities---visual, textual, and auditory---were obtained. In order to fully reveal the influence of each of the modalities, independent ground-truth memorability scores are required to elucidate the role they each play when coinciding with one another in a multi-modal medium such as video, and should be a focus of future memorability tasks and research.

\begin{table}[t]
\caption{Official results on test-set for sub-task 1 for the TRECVid dataset.}
\label{tab:prediction1}
\hspace*{-0.2cm}
\begin{tabular}{ccccc}
    \toprule
     &\multicolumn{2}{c}{Short-norm} &\multicolumn{2}{c}{Long}\\\cline{2-3}\cline{4-5}
    \textbf{Run} & \textbf{r\textsubscript{s}} & r & \textbf{r\textsubscript{s}} & r \\
    BayesianRidge Dense121 & 0.053 & 0.071 & - & -\\
    xResNet50 Transfer Frames &  0.105 & 0.13 & -.021 & -.036\\
    AWD-LSTM Transfer Caption &  0.105 & 0.083 & 0.002 & 0.013\\
    \bottomrule
\end{tabular}
\end{table}

\begin{table}[t]
\caption{Official results on test-set for sub-task 1 for the Memento10k dataset.}
\label{tab:prediction2}
\hspace*{-0.2cm}
\begin{tabular}{cccccccc}
    \toprule
     &\multicolumn{2}{c}{Short-raw} &\multicolumn{2}{c}{Short-norm}\\\cline{2-5}
    \textbf{Run} & \textbf{r\textsubscript{s}} & r & \textbf{r\textsubscript{s}} & r\\
    BayesianRidge Dense121 & \textbf{0.523} & 0.522  & \textbf{0.524} & 0.522\\
    BayesianRidge Vggish & 0.29 & 0.289  & 0.272 & 0.274\\
    AWD-LSTM Caption & - & -  & 0.174 & 0.172\\
    AWD-LSTM Transfer Caption & - & -  & 0.181 & 0.163\\
    xResNet50 Frames & - & -& 0.129 & 0.114\\
    \bottomrule
\end{tabular}
\end{table}

\begin{table}[t]
\caption{Official results on test-set for sub-task 2 for the TRECVid dataset.}
\label{tab:generalisation}
\hspace*{-0.2cm}
\begin{tabular}{cccccccc}
    \toprule
     &\multicolumn{2}{c}{Short-raw} &\multicolumn{2}{c}{Short-norm}\\\cline{2-3}\cline{4-5}
    \textbf{Run} & \textbf{r\textsubscript{s}} & r & \textbf{r\textsubscript{s}} & r\\
    BayesianRidge Vggish Memento & 0.018 & 0.008  & 0.021 & 0.012\\
    xResNet50 Frames LaMem & 0.093 & 0.073  & 0.088 & 0.076\\
    xResNet50 Frames Memento & \textbf{0.116} & 0.131  & \textbf{0.132} & 0.145\\
    AWD-LSTM Caption Memento & 0.114 & 0.12 & 0.106 & 0.11\\
    \bottomrule
\end{tabular}
\end{table}

\section*{Acknowledgements}
This work was funded by Science Foundation Ireland (SFI) under Grant Number SFI/12/RC/2289\_P2, co-funded by the European Regional Development Fund.

\bibliographystyle{ACM-Reference-Format}
\def\bibfont{\small} 
\bibliography{sigproc}
\end{document}